\documentclass[journal]{IEEEtran}

\usepackage[T1]{fontenc} 
\usepackage{amsmath,amssymb}
\usepackage{txfonts}
\usepackage{bm} 

\usepackage[compress,noadjust]{cite}

\usepackage{hyperref}
\usepackage{graphicx}
\usepackage{placeins}
\usepackage{float}
\usepackage{booktabs}
\usepackage{color}

\usepackage[textwidth=1.5cm]{todonotes}

\usepackage{chngcntr}
\counterwithin{paragraph}{section}

\def\vector#1{{\boldsymbol{#1}}}

\begin{document}
\bstctlcite{IEEEexample:BSTcontrol}

\author{Julius~Hannink, Thomas~Kautz, Cristian~F.~Pasluosta, Jens~Barth, Samuel~Sch\"ulein, Karl-G\"unter Ga\ss mann, Jochen~Klucken,  Bjoern~M.~Eskofier,~\IEEEmembership{Senior Member,~IEEE,~EMBS}
	
	\thanks{J. Hannink, T. Kautz, J. Barth and B. M. Eskofier are with the Digital Sports Group, Department of Computer Science, University of Erlangen-N\"urnberg (FAU), Germany. C. F. Pasluosta was with the Digital Sports Group (FAU) and recently joined the Laboratory for Biomedical Microtechnology, Department of Microsystems Engineering, University of Freiburg, Germany.}
	\thanks{S. Sch\"ulein, K.-G. Ga\ss mann are with the Geriatrics Centre Erlangen, Waldkrankenhaus St. Marien, Erlangen, Germany.}
	\thanks{J. Klucken is with the Department of Molecular Neurology, University Hospital Erlangen, University of Erlangen-N\"urnberg (FAU), Germany.}
	\thanks{Corresponding author: J. Hannink, \href{mailto:julius.hannink@fau.de}{julius.hannink@fau.de}}
}

\title{Stride Length Estimation with Deep Learning}

\maketitle

\begin{abstract}

\textit{Objective}: Accurate estimation of spatial gait characteristics is critical to assess motor impairments resulting from neurological or musculoskeletal disease. Currently, however, methodological constraints limit clinical applicability of state-of-the-art double integration approaches to gait patterns with a clear zero-velocity phase.
\textit{Methods}: We describe a novel approach to stride length estimation that uses deep convolutional neural networks to map stride-specific inertial sensor data to the resulting stride length. The model is trained on a publicly available and clinically relevant benchmark dataset consisting of 1220 strides from 101 geriatric patients.  
Evaluation is done in a 10-fold cross validation and for three different stride definitions.

\textit{Results}: Even though best results are achieved with strides defined from mid-stance to mid-stance with average accuracy and precision of $\boldsymbol{0.01 \pm 5.37\,\text{cm}}$, performance does not strongly depend on stride definition. The achieved precision outperforms state-of-the-art methods evaluated on the same benchmark dataset by $\boldsymbol{ 3.0\,\text{cm }(36\%)}$. 
\textit{Conclusion}: Due to the independence of stride definition, the proposed method is not subject to the methodological constrains that limit applicability of state-of-the-art double integration methods.  Furthermore, it was possible to improve precision on the benchmark dataset. 
\textit{Significance}: With more precise mobile stride length estimation, new insights to the progression of neurological disease or early indications might be gained. Due to the independence of stride definition, previously uncharted diseases in terms of mobile gait analysis can now be investigated by re-training and applying the proposed method.

\end{abstract}
\begin{IEEEkeywords}
deep learning, convolutional neural networks, mobile gait analysis, stride length, regression
\end{IEEEkeywords}

\section{Introduction}

In a variety of neurological or musculoskeletal diseases, motor impairment leads to specific gait characteristics. In Parkinson's disease (PD), for example, the resulting symptoms of gait impairment are often characterised by shuffling steps, reduced stride length, or impaired gait initiation \cite{Klucken2013}. In advanced stages, PD patients can even experience spontaneous episodes of gait freezing. These symptoms can lead to a severe reduction in patient mobility and quality of life \cite{Ellis2011} and it is thus important to quantify and treat gait impairments as early as possible. However, the physician's rating of gait quality e.g. in terms of the Movement Disorder Society's revision of the Unified Parkinson Disease Rating Scale (MDS-UPDRS) \cite{Goetz2008} is often biased by subjectiveness.   
Moreover, it requires long-time experience by the movement disorder specialist, is thus costly and restricted to the availability of the physician. With an ageing world population (\cite{UN2015,EU2005}) and a resulting increase in prevalence of chronic age-related disorders such as PD, this also requires a change in the healthcare system towards a new era of diagnostics and individualized treatment \cite{Pasluosta2015}.   
  

In the past, this has led to the development of several electronic measurement systems with the aim to extract spatio-temporal gait parameters and aid the physician in an objective gait impairment scoring. These systems include for example camera based, stationary 3D optical tracking of markers affixed to the body \cite{Kressig2004} or electronic walkways with embedded pressure sensors \cite{Givon2009}. Optical tracking systems require time-consuming preparation of the subject, but enable three-dimensional, sub millimetre accurate tracking. Computerized walkways with embedded pressure sensors, however, require no preparation of the subject, but are limited in accuracy. Furthermore, they can only assess gait parameters related to ground contact, i.e. foot angles outside the ground plane can not be tracked. Both of these systems are expensive and need a laboratory environment for gait analysis which prohibits their use in everyday clinical practice or outpatient monitoring.

Unlike stationary systems, inertial sensors placed at the lower extremity enable mobile and unobtrusive gait analysis that is unconstrained by the laboratory setting \cite{Klucken2013}. This type of measurement system is inexpensive and requires little subject-preparation while at the same time enabling estimation of three-dimensional spatio-temporal gait parameters (\cite{Mariani2010,Rampp2015,Aminian2002,Salarian2004,Rebula2013,Trojaniello2014,Ferrari2015}). Although mobile systems are increasingly developed to support clinical diagnostics, monitor medical treatment or provide early indicators for gait impairment, they are still an open field of research \cite{Pasluosta2015}. One of the most crucial components of such systems is accurate estimation of stride-by-stride parameters. Therefore, the purpose of this work is accurate estimation of stride length based on stride-specific inertial sensor data captured at the subject's feet.

\section{Related Work}
There is a growing body of literature regarding stride length estimation from inertial sensors placed at the lower extremity of the human body. The methods covered generally divide into two classes: (Bio)mechanical model based approaches (\cite{Aminian2002,Salarian2004,Salarian2013}) and double-integration approaches (\cite{Veltink2003,Sabatini2005,Mariani2010,Rebula2013,Trojaniello2014,Rampp2015,Ferrari2015}).
\\
\paragraph{Model based approaches}
The model based approaches employ a double pendulum model for the swing phase and an inverse double pendulum for the stance phase. Three joints (hip and knees, ankle is considered fixed) and four segments (shank and thigh) are modelled. Integration of the respective equations of motion is driven by gyroscope data captured at the shank and thigh of the subjects (\cite{Aminian2002,Salarian2013,Salarian2004}). Salarian et al. \cite{Salarian2013} additionally try to reduce the number of sensing units either by duplicating data of the right thigh on the left thigh or by simplifying the model to a single-pendulum with two sensors at the shanks.
Apart from the amount of sensors needed on the patient, the models described in \cite{Aminian2002,Salarian2004} and \cite{Salarian2013} confine the movement-model to the sagittal plane and thus limit their applicability for analysis of impaired gait. Furthermore, specific subject characteristics (segment length etc.) are needed to scale the biomechanical models. One of the major benefits, however, is the intrinsic implementation of biomechanical constraints between the two legs.
\\
\paragraph{Double-integration approaches}
The vast majority of stride length estimation algorithms comes from the class of double-integration methods. This family of algorithms makes use of the stance phase as a zero-velocity update point (ZUPT) to re-initialize the integration process. Due to intrinsic measurement errors in currently available inertial sensors, this is necessary in order to minimize the integration drift \cite{Peruzzi2011}. For one gait cycle, the algorithm consists of sensor orientation estimation from gyroscope measurements in order to convert sensor readings from the local, constantly changing sensor frame to global coordinates. This is followed by gravity cancellation and double-integration of accelerometer readings. Due to aforementioned limitations in currently available mobile sensors, the integration result has to be corrected or de-drifted in a last processing step by enforcing several constrains (e.g. zero-velocity at the start and end of the stride, level floor, etc.). This class of methods is mostly evaluated per leg and biomechanical constrains are hard to implement. However, three-dimensional sensor trajectories can be reconstructed allowing a deeper analysis of gait characteristics or disease-specific alterations.

The contributions on the double-integration approach covered here (\cite{Veltink2003,Sabatini2005,Mariani2010,Rebula2013,Trojaniello2014,Rampp2015,Ferrari2015}) all employ accelerometer and gyroscope data captured at the subjects' feet, Trojaniello et al. \cite{Trojaniello2014} additionally use magnetometer information. Sampling rates are comparable and between 100 and 200 Hz for all studies. The contributions mainly differ in the choice of orientation estimation method. Rampp et al. \cite{Rampp2015} and Mariani et al. \cite{Mariani2010} apply a quaternion-based time integration of angular rates to estimate the sensor orientation, as proposed by Sabatini \cite{Sabatini2005}. Other contributions (\cite{Rebula2013,Trojaniello2014,Ferrari2015}) make use of a Kalman filter to obtain sensor orientations. Integration is mainly done directly, i.e. forward in time. Only Trojaniello et al. \cite{Trojaniello2014} use a direct and reverse integration scheme which spawns two integrations -- one from the start and one from the end of the current stride -- and computes a weighted mean between the two with weights depending on the distance to the next ZUPT. This is said to further resolve integration drifts \cite{Zok2004}.
De-drifting of integrated gravity-free acceleration signals is done in the spatial domain in \cite{Rebula2013,Rampp2015} and \cite{Mariani2010} with a linear de-drifting function.
Trojaniello et al. \cite{Trojaniello2014} use a frequency-based de-drifting approach to suppress the low-frequency drift component in the integrated signals. Another possibility is to combine all steps above in one Kalman filter  estimating orientation, velocity and position from angular rates and acceleration. This is implemented by Ferrari et al. \cite{Ferrari2015}.
\\
\paragraph{Clinical applicability}
Tab. \ref{tab:relatedwork} summarizes the dataset characteristics and results on all related work covered here.
Regarding the mean accuracies and precisions, the majority of methods are similar with accuracies within $1$ cm and precisions around $6 - 8$ cm. There are two exceptions: Rebula et al. \cite{Rebula2013} reach good precision but are outperformed by others in terms of accuracy. The method proposed by Trojaniello et al. \cite{Trojaniello2014} achieves the best accuracy and precision almost resolving the reference precision of $\pm 1.27\,\text{cm}$ \cite{Trojaniello2014}. They are the only ones employing a magnetometer which might greatly improve their orientation estimation. However, this is only true if indoor magnetic field distortions by e.g. ferroconcrete in the proximity of the gait acquisition track are small enough. In a general setting without prior knowledge about earth magnetic field disturbances due to construction materials, this out-rules the use of magnetometers for indoor gait data acquisitions with wearable sensors \cite{DeVries2009}.

This also applies to other studies listed in Tab. \ref{tab:relatedwork}.
Regarding the number and type of subjects, there are major differences in terms of variability within the datasets. Not listed are other quite important characteristics of the study populations, i.e. stride length distribution, diversity of gait alterations, precise age distribution, gender, study protocol or track length used during data acquisition. This is greatly threatening the comparability of methods and their clinical application due to non-representative study populations. It shows the urgent need of a unified evaluation of all stride length estimation methods on the same, clinically relevant benchmark dataset.  



\newlength{\doubledaggerlength}
\settowidth{\doubledaggerlength}{\footnotesize$^{\dagger\dagger}$}
\def\doubledaggerspace{\hspace{\doubledaggerlength}\null}
\begin{table*}[!ht]
	\centering
	\caption{Overview of related work: Dataset characteristics vs. mean accuracies and precisions}
	\begin{tabular}{lrlrlrr}
		\toprule
		& Subjects & Diagnose &\# Strides & Reference &  Mean acc. $\pm$ prec. & rel. prec. \\\toprule
		Rebula et al. \cite{Rebula2013} 			& 9 \hspace{.2em}young 	& healthy		&5538	& MoCap	&$-1.2\pm3.7\text{ cm}^{\dagger\dagger}$ & $3.2~ \%^{~}$ \\
		Mariani et al. \cite{Mariani2010}			& 10 \hspace{.2em}young		& healthy		&482	& MoCap	&$2.4\pm7.5$ cm\doubledaggerspace& $6.8~ \%^{~}$\\\midrule
		Mariani et al. \cite{Mariani2010}	  		& 10 elderly 	& healthy		&492	& MoCap	&$0.7\pm6.1$ cm\doubledaggerspace& $6.1~ \%^{~}$\\
		Trojaniello et al. \cite{Trojaniello2014} 	& 10 elderly 	& healthy		&576	& GaitRite	&$0.0\pm 1.9$ cm\doubledaggerspace& $1.0~ \%^{~}$ \\
		Trojaniello et al. \cite{Trojaniello2014} 	& 10 elderly 	& Parkinson's	&532	& GaitRite	&$0.1\pm 1.9$ cm\doubledaggerspace& $2.0~ \%^{~}$ \\
		Rampp et al. \cite{Rampp2015} 				& 101 elderly 	& geriatric 	&1220	& GaitRite	&	 $-0.3\pm8.4$ cm\doubledaggerspace& $10.7~ \%^\dagger$\\
		Ferrari et al. \cite{Ferrari2015} 			& 14 elderly 	& Parkinson's 	&1314	& GaitRite	&	 $-0.2\pm7.0$ cm\doubledaggerspace& $5.0~ \%^{~}$\\
		Salarian et al. \cite{Salarian2013} 			& 15 elderly 	& healthy \& impaired 	&229	& MoCap	&	 $-0.8\pm6.6$ cm\doubledaggerspace& $6.0~ \%^{\dagger}$\\
		\bottomrule
		\multicolumn{7}{l}{
		\footnotesize $^\dagger$ ~Computed based on mean stride length reported.\hfill 
		\footnotesize $^{\dagger\dagger}$ Computed after correspondence with the authors.}
	\end{tabular}	
	\label{tab:relatedwork}
\end{table*}

Although there has been much progress on stride length estimation from inertial sensor data over the last years, a variety of clinically relevant phenomena still cannot be resolved due to the achieved measurement precision. These phenomena include for example the decline in stride length with age in healthy controls, the reduction in stride length with disease progression in e.g. PD patients or the effect of medication. Hollman et al. \cite{Hollman2011} report normative data on mean stride length in elderly, healthy controls. Between the age groups 70-75 years and 85+ years, stride length decreases by 20 cm in males and 14 cm in females in the mean values over 294 participants \cite{Hollman2011}. This translates to a reduction around 1-2 cm/year. Resolving this difference with mobile sensors requires the effect to be larger than typically twice the standard deviation of the error distribution (i.e. precision) obtained in the validation study of a stride length estimation method. In order to resolve the yearly decline with age, the precision of a mobile stride length estimation method would therefore need to be around 5 mm.   
A second clinical application involves stride length estimation in PD patients. Hass et al. \cite{Hass2012} report normative values on mean stride length in PD patients over the course of the disease as measured by the Hoehn\&Yahr (H\&Y) scale \cite{Hoehn1967}. Between the groups H\&Y < 1.5 (mild) and H\&Y = 3-4 (severe), they observed a reduction in mean stride length of 11 cm for males and 12 cm for females based on data from 310 PD patients \cite{Hass2012}. Following the argument above, mobile stride length estimation methods would require precisions around 5 cm in order to resolve the effect of stride length reduction with disease progression even on the very coarse H\&Y scale ranging from 0-5. Lastly, Bryant et al. \cite{Bryant2011} investigated the change in stride length between the best possible (on) and worst (off) medication state in PD patients. Based on data from 21 PD patients, they found a stride length difference of 18 cm between the on- and off-state. These values help to get a feeling about the ranges of stride length effects between best possible to worst medication levels or mild to severe disease progression. Moreover, the clinician is probably interested in more minute changes in clinical scores or medication levels corresponding to smaller effects on the parameter stride length. To this date, mobile stride length estimation is not yet precise enough for this kind of tasks. From a clinical perspective, the problem thus stays challenging.
\\
\paragraph{Towards a new type of methodology}
Until now, the choice of method for stride length estimation has been based on biomechanical, physical or geometrical reasoning and these approaches have proven to work in technical validation settings. However, one could also aim to learn a regression function linking the raw sensor data of a stride directly with the corresponding stride length. Similar approaches were reported by Aminian et al. \cite{Aminian1995} in 1995 where a two-layer perceptron was used to estimate speed and incline of human walking. Due to the limits on computing power 20 years ago, the neural network structure presented by Aminian et al. \cite{Aminian1995} is very shallow with 10 input nodes, one hidden layer of five units and one output node. A 10 feature parametrization of the accelerometer signal based on physiological and statistical reasoning is used as input to the network.

With the recent advances in deep learning (\cite{LeCun2015,Krizhevsky2012,Hinton2006}) and user-friendly interfaces to such techniques like torch \cite{torch7} or google's TensorFlow library \cite{tensorflow2015-whitepaper}, the parametrization of the input data has become obsolete. By regressing against raw sensor data instead of parametrizations and by employing sufficiently deep architectures, the true potential of the neural network approach can be utilized. Based on a publicly available and clinically relevant benchmark dataset, we describe a method to learn a non-linear relationship between stride-specific sensor data and the corresponding stride length with a convolutional neural network. The network can be trained on big data sets \cite{Chen2014} while at the same time promising excellent individualisation capabilities to the patient at hand \cite{Stober2015}. 
We evaluate the approach in a cross-validation and obtain mean accuracy and precision regarding stride length estimation. In order to assess the method's dependency on the data segment provided as input, this is done for three different stride definitions. These are either based on biomechanical events (heel-strike and mid-stance) or contain the relevant information (i.e. the swing phase where stride length is produced) but are not directly related to biomechanical events in the datastream.

The method we propose here adds a completely novel approach to the problem of stride length estimation from inertial sensor data. It is not subject to the zero-velocity assumption that the double-integration approaches need to re-initialize the integration process. Thereby, the current work enables spatial gait parameter estimation in situations where this assumption is violated, e.g. in patients with spasticity. Compared to the long processing chain employed by double-integration methods that results in amplification of errors and requires cumbersome, careful and error-prone tuning of counter measures such as de-drifting, the method proposed here is purely data-driven. There is no need to carefully keep track of and account for the accumulating error during processing.

Our main contributions are: 1) We show feasibility regarding estimation of spatial gait characteristics with deep convolutional neural networks. 2) With that change in methodology, we drop the zero-velocity phase assumption that constrains state-of-the-art double integration methods and limits their applicability.

\section{Methods}
\begin{figure}[!h]
		\centering
		\includegraphics[width=.6\columnwidth]{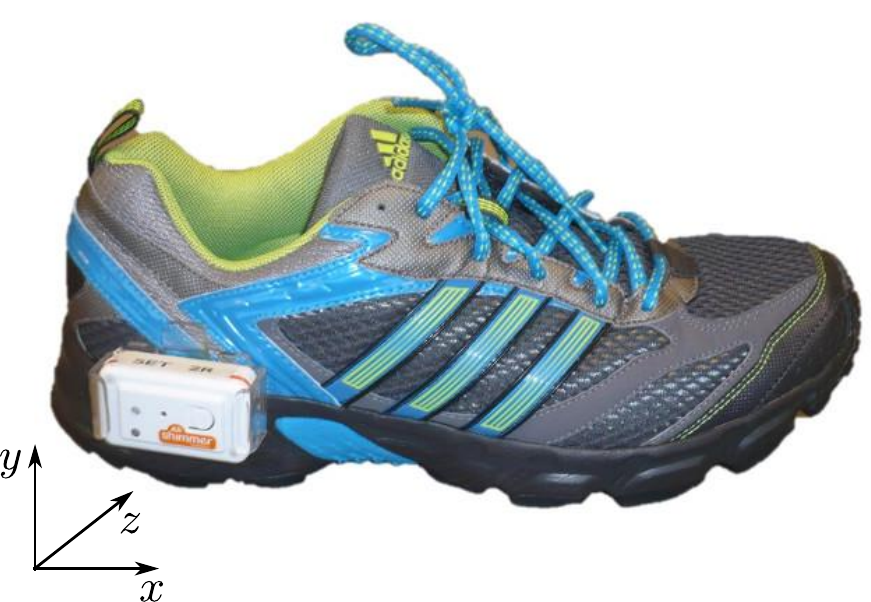}
		\caption{Placement of the inertial sensor and axes definition.}
		\label{fig:shoes}
\end{figure}

\paragraph{Data Collection and Setup}
We use a benchmark dataset collected by Rampp et al. \cite{Rampp2015} that is publicly available at \url{https://www5.cs.fau.de/activitynet/benchmark-datasets/digital-biobank/} and briefly described here. 
For data collection, the Shimmer2R sensor platform \cite{Burns2010} was used. It contains a 3d-accelerometer (range $\pm 6\,\text{g}$) and a 3d-gyroscope (range $\pm 500\, ^\circ$/s) and was attached laterally right below each ankle joint (see Fig. \ref{fig:shoes}). The same shoe model (adidas Duramo 3) was used by all subjects to avoid gait changes due to different shoe characteristics \cite{Menant2009}. Data was captured at $102.4$ Hz and a resolution of $12$ bit. Simultaneously, validation data was acquired with the well established, instrumented pressure mat GAITRite (\cite{McDonough2001,Webster2005}). The sensitive area of the pressure mat was $609.60~ \text{cm} \times 60.96~ \text{cm}$ with a spatial resolution of $\pm 1.27$ cm and a track width of $84$ cm.

In total, 116 geriatric inpatients were assessed with this setup at the Geriatrics Centre Erlangen (Waldkrankenhaus St. Marien, Erlangen, Germany). Written informed consent was obtained prior to the gait assessment in accordance with the ethical committee of the medical faculty at Friedrich-Alexander University Erlangen-N\"urnberg (Re.-No. 4208).

Patients performed a detailed geriatric assessment, described in detail by Rampp et al. \cite{Rampp2015}. For the scope of this manuscript, we focus on the free walking test over $10\,\text{m}$ at comfortable walking speed instrumented with the inertial sensors and the GAITRite system. Here, the number of strides acquired from each patient ranges from 7 -- 32 with a mean of 12 strides per patient. After excluding datasets from eight patients due to medical reasons, two due to inertial sensor malfunction and additional five due to measurement errors with the GAITRite system, a total of 101 datasets were left for training and evaluation of the stride length estimation method proposed here. In 54\% of the study population, gait disorders or fall proneness were diagnosed. The other top three diagnoses were heart rhythm disorder (70\%), arterial hypertension (69\%) and coronary artery disease (41\%) all of which are associated with gait and balance disorders \cite{Salzman2010}. Therefore, this dataset constitutes a clinically relevant study population both in terms of the number of subjects and the presence of unpredictable gait alterations.
\\
\paragraph{Preprocessing}
Before the inertial sensor data is fed to the convolutional neural network, we perform a series of preprocessing steps. These include extraction of annotated strides (a segmentation of the continuous signal into strides is already supplied by the dataset) and calibration from raw sensor readings to physical units \cite{Ferraris1995}. Due to different sensor mounting on the shoe, coordinate system transformations are needed to align sensor axes on left and right feet. Furthermore, the signals from accelerometer and gyroscope are normalised w.r.t. the respective sensor ranges and zero-padded to fixed length of 256 samples per stride to ensure equally scaled and fixed size input to the network.
\begin{figure}[h]
		\centering
		\includegraphics[width=.95\columnwidth]{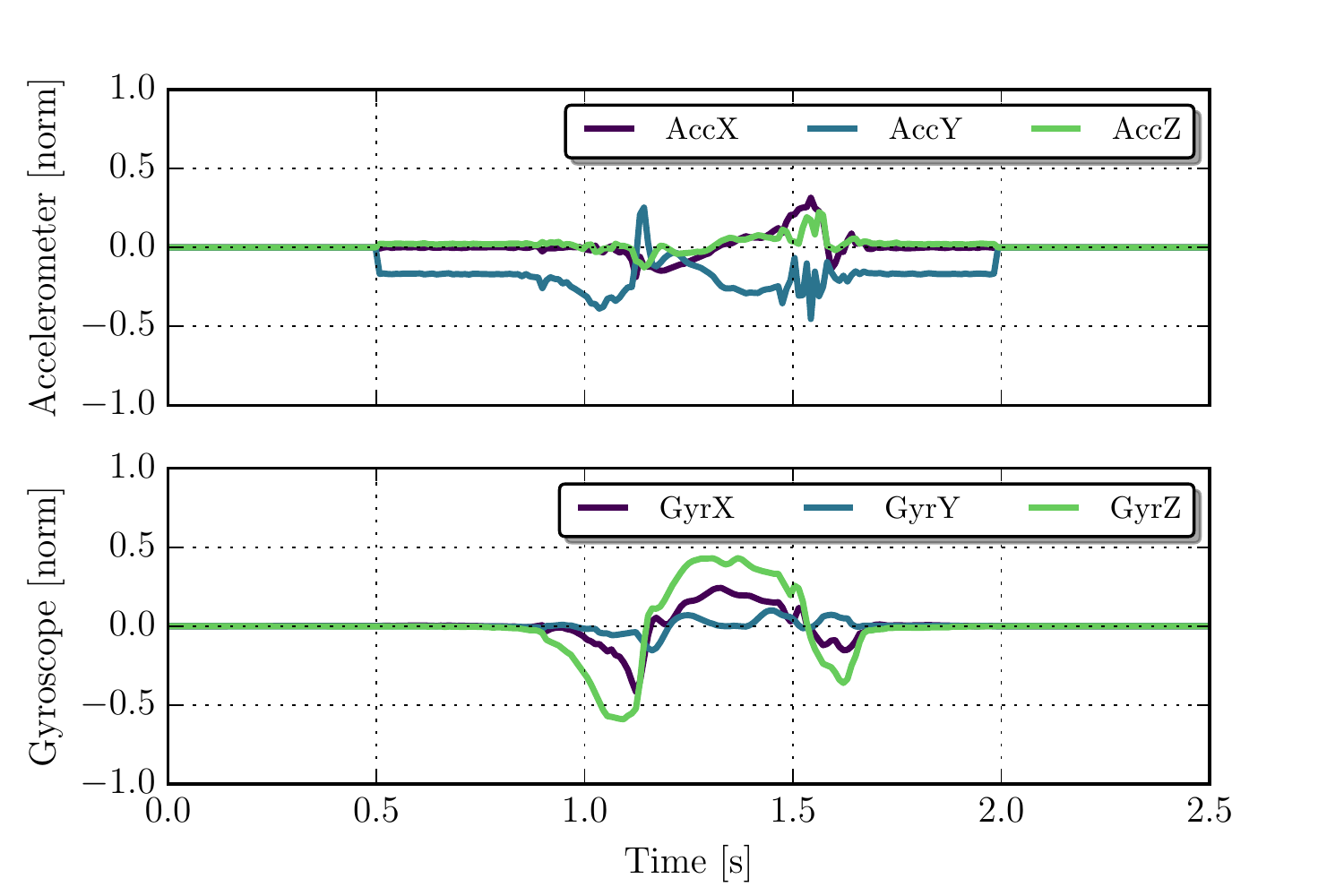}
		\caption{Exemplary input signal for a stride defined from MS$\to$MS after preprocessing.}
		\label{fig:inputSignal}
\end{figure}

Since we will compare the algorithm on different stride definitions, we need to detect mid-stance (MS) and heel-strike (HS) events in the stride segmentation provided by the dataset and adjust the stride borders accordingly. Detection of these two events within the sensor data is done according to \cite{Rampp2015}. We call the stride definition provided by the dataset msDTW because of the multi-dimensional subsequence dynamic time warping they used for segmentation (for details see \cite{Barth2015,Rampp2015}).

An exemplary, pre-processed input signal for the network for one stride defined from MS$\to$MS is shown in Fig. \ref{fig:inputSignal}.\\

\paragraph{Network Architecture}

\begin{figure}[!h]
\centering
\includegraphics[width=.9\columnwidth]{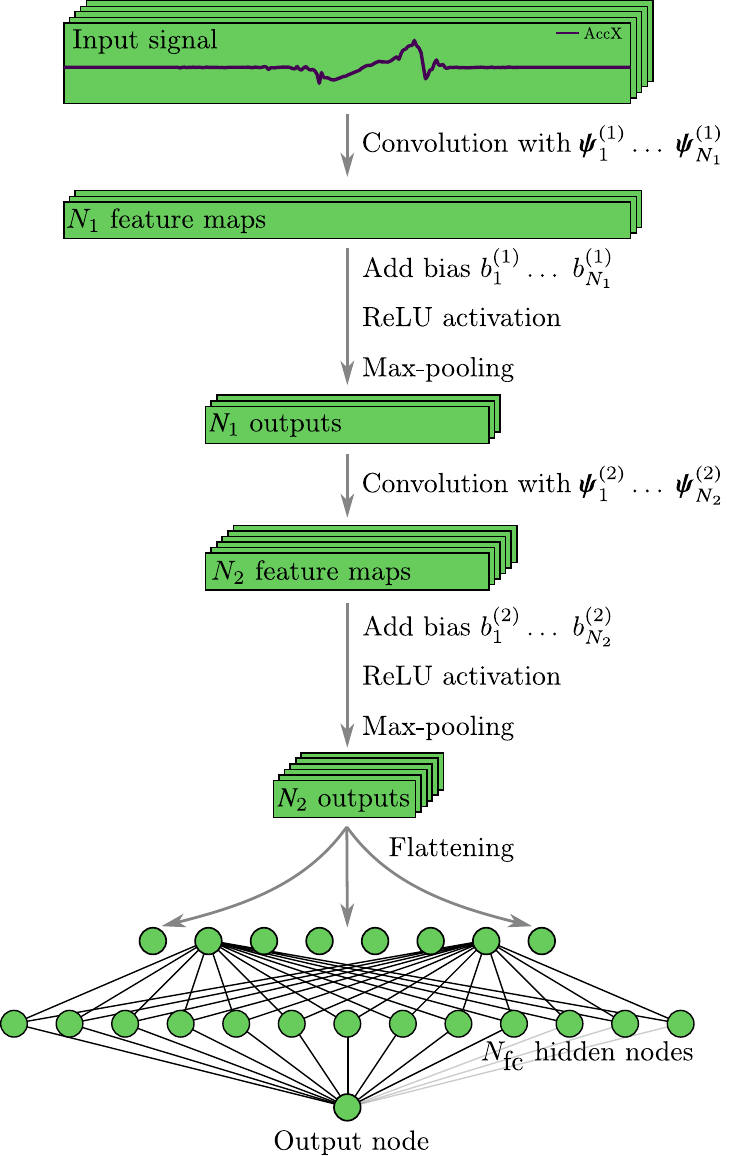}
\caption{Architecture for the neural network regression approach presented in this paper: Two convolutional layers followed by max-pooling, one fully connected layer and a readout layer contracting the last hidden layer to the single target variable.}
\label{fig:NetArchitecture}
\end{figure}

The neural network architecture we choose is a two layer convolutional network followed by one fully connected layer and a readout-layer (see Fig. \ref{fig:NetArchitecture}). The main theoretical motivation for this choice is the locality of features gained by this type of architecture \cite{LeCun2015}. The network is thus forced to take the true topology of the input data as multi-channelled, synchronized time series into account and maintain its temporal context. It only considers local connections and does not link	 information from temporally far away regions in the input signal.

The preprocessed inertial sensor data $x_{i,j}$ with $i=1\dots L_0$ samples and $j = 1\dots N_0$ channels serves as input to the network. In the first convolutional layer, feature maps are constructed from $x_{i,j}$ by convolution with $k=1\dots N_{1}$ kernels $\psi^{(1)}_{i,j,k}$ of size $i=1\dots L_1$ and $j\in[1,N_0]$. A bias term $b^{(1)}_k $ is added to each feature map before a ReLU activation function \cite{Glorot2011} produces the output activations of the first convolutional layer
\begin{equation}
	\vector c^{(1)}_{k} = \text{ReLU}\left(
						\sum_j\vector\psi^{(1)}_{j,k} * \vector x_{j} +b^{(1)}_{k}
						\right)
	\label{eq:outputNodesConv1}
\end{equation}
with $k\in[1,N_1]$ and $\text{ReLU}(z) = \max(0,z)$.

After each convolutional layer, we insert a max-pooling layer that downsamples the feature maps by taking the maximum in non-overlapping windows of length $r$. The downsampling factor thus is $1/r$. The output from the first max-pooling layer is called $o^{(1)}_{i,j}$ with $i\in [1,L_0/r]\text{ and } j\in[1,N_1]$.

In the second convolutional layer, the same approach is repeated with different kernels and biases. Given the input signal $o^{(1)}_{i,j}$, feature maps are constructed by convolution with $k=1\dots N_2$ kernels $\psi^{(2)}_{i,j,k}$ of length $i=1\dots L_2$ and $j\in[1,N_1]$. A bias term $b^{(2)}_k$ is added before a ReLU activation function produces the output nodes 
\begin{equation}
	\vector c^{(2)}_{k} = {ReLU}\left(
						\sum_j\vector\psi^{(2)}_{j,k} * \vector o^{(1)}_j +b^{(2)}_{k}
						\right)
	\label{eq:outputNodesConv2}
\end{equation}
with $k\in[1,N_2]$. After max-pooling, the output $o^{(2)}_{i,j}$ is of size $i\in [1,L_0/r^2]\text{ and } j\in[1,N_2]$. 
Similarly, up to $\log_{r} L_0$  convolutional layers followed by max-pooling could be stacked before the input signal is downsampled to size 1.

For the fully connected layer, the output from the last max-pooling layer $o^{(2)}_{i,j}$ is flattened to a one dimensional vector $\tilde o^{(2)}_i$ with $i\in[1, N_2\,L_0/r^2]$. This vector is multiplied with the  weights $W^\text{(fc)}_{i,j}$ with $i=1\dots(N_2\,L_0/r^2)$ and $j=1\dots N_\text{fc}$. A bias term $b^\text{(fc)}_j$ is added before a ReLU activation function produces the output of the fully connected layer
\begin{equation}
	o^\text{(fc)}_j = \text{ReLU}\left(\sum_i W^\text{(fc)}_{i,j} \; \tilde o^{(2)}_{i}  + b^\text{(fc)}_j\right)
	\label{eq:outputNodesHidden}
\end{equation}
with $j\in[1,N_\text{fc}]$.

Finally, a readout layer compresses the $N_\text{fc}$ output nodes of the hidden layer to the single target variable stride length. This is done by multiplication with a weight vector $W^\text{(ro)}_{i,1}$ with $i \in [1,N_\text{fc}]$. A final bias $b^\text{(ro)}$ is added to arrive at the estimate of the target variable
\begin{equation}
	y = \sum_i W^\text{(ro)}_{i,1} o^\text{(fc)}_i + b^\text{(ro)}
	\label{eq:target estimate}
\end{equation}
Similarly, a target vector $(y_1, \dots ,y_n)$ representing a set of stride features could be estimated by changing the dimensionality of weight and bias in the readout layer. 
 
For our application of stride length estimation, the input data is of length $ L_0 = 256$ with $N_0 = 6$ channels. On the first layer of the network, we chose to learn $N_1=32$ filters of length $L_1=30$ samples, followed by $N_2=64$ and $L_2 = 15$ on the second. 
Max-pooling is done in non-overlapping windows of length $r = 2$ and the fully connected layer has $N_\text{fc} =1024$ nodes. Given the downsampling factor and the sampling rate, the filter lengths both correspond to approximately $0.29$ s. The theoretical motivation for this choice is to keep the resulting receptive field size constant ($L_1 = r \, L_2 $) while increasing the amount of features maps ($N_2 > N_1$) with network depth.\\

\paragraph{Learning a Regression Function}

Given a training dataset, we now focus on determining the parameters involved in the previously described network architecture, i.e. kernels/weights and biases on each layer. 

Training neural networks is commonly posed as an optimization problem regarding a scalar error function (implicitly) depending on the network parameters. This error implements a discrepancy measure between the predicted output and a ground truth reference on the training dataset or subsets thereof. Using back-propagation, weights and biases on all layers are changed with the aim to minimize the error. In practice, only random subsets of the training dataset, called mini-batches, are shown to the optimizer in one iteration of the training loop to speed up the learning phase (stochastic learning). 

Given the relative error distribution 
\begin{equation}
\epsilon_i = \frac{y_i-y_{i,\text{ref}}}{y_{i,\text{ref}}} \;\;\text{with }i\in[1,N_\text{batch}]
\label{eq:relativeError}
\end{equation}
on such a mini-batch of size $N_\text{batch}$, we define the discrepancy measure to minimize as
$E(\vector\epsilon) = \text{rmsq}(\vector\epsilon)$.
The term $\text{rmsq}(\vector\epsilon)$ is the root-mean-square-error on the mini-batch.


For optimization, we use Adam \cite{Kingma2015} which is a state-of-the-art optimization method for stochastic learning. It shows faster convergence than other stochastic optimization routines on benchmark datasets and we use default settings of $\alpha=1\text{e}^{-3},\beta_1=0.9,\beta_2 = 0.999 \text{ and } \epsilon=1\text{e}^{-8}$ (for details see \cite{Kingma2015}). All weights are initialized randomly by sampling a truncated normal distribution with standard deviation 0.1 and biases are initialized from 0.1. Training is done for a fixed number of 4000 iterations with a mini-batch size $N_\text{batch}=100$ strides.

As a measure against over-fitting we use dropout on the fully connected layer. Dropout effectively samples a large number of thinned architectures on the hidden layer by randomly dropping nodes and their connections during training. This technique has been shown to prevent over-fitting significantly in many use-cases and is superior to weight-regularisation methods \cite{Srivastava2014}. We use a fixed dropout probability of $p_\text{drop} = 0.5$, so every node on the fully connected layer has a 50/50 chance of being dropped during training. During testing, however, the full architecture is used and no connections are dropped.

The network is implemented and trained using google's TensorFlow library \cite{tensorflow2015-whitepaper}.
\\

\paragraph{Evaluation Scheme}
Evaluation of the proposed method is based on a 10-fold cross validation scheme. The strides from 101 patients on the dataset are sorted into training and test set depending on the patient identifier to ensure distinct splits of the dataset.
For each of the three stride definitions msDTW, HS$\to$HS and MS$\to$MS, the model is evaluated in such a cross-validation scheme and the stride length is estimated on the test set in each fold. The predictions from individual folds are then pooled to arrive at average statistics for the current stride definition. 
Evaluation statistics include average accuracy $\pm$ precision which correspond to the mean and standard deviation of the (signed) error distribution. Additionally, we state precision relative to the mean stride length, average absolute accuracy $\pm$ precision as well as Spearman correlation coefficients between the reference system and our approach. 

In order to assess the learning speed and performance of the convolutional neural network, we compute the training error and the corresponding precision over the training iterations for an exemplary 90/10\% (patient-wise) split of the dataset.

\section{Results}

\paragraph{Training}
On a mobile workstation equipped with an Intel Core i7 processor (8 cores), 16GB of RAM and no GPU acceleration, training takes around 20 min. per fold.

\begin{figure}[!h]
		\centering
		\includegraphics[width=.95\columnwidth]{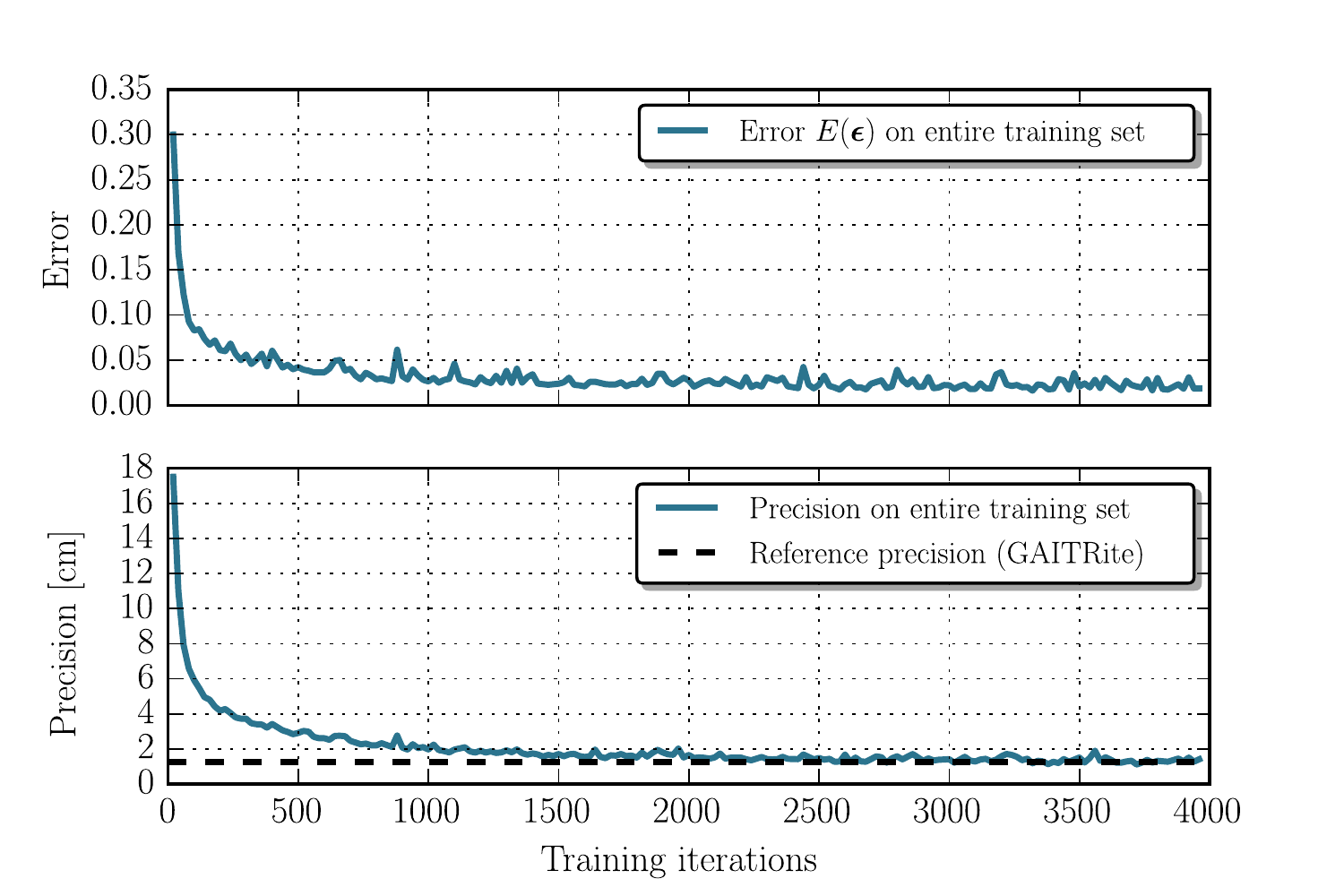}
		\caption{Error $E(\vector\epsilon)$ and precision evaluated on the entire training set for an exemplary 90/10\% train/test split of the dataset during training.}
		\label{fig:training}
\end{figure}

For an exemplary fold, Fig. \ref{fig:training} shows the error as well as the precision on the entire training set over the iterations. The fixed number of 4000 iterations is sufficient to reach a stable regime of the error $E(\vector\epsilon)$ and precision evaluated on the entire training dataset. Moreover, we reach the reference precision after training which demonstrates the usability of the described approach for estimating stride length at high precision with a model tuned to the individual. Because we can not formally evaluate the performance of individualized models here due to the lack of data (6 strides on average per foot and patient), the results on the training set serve as a substitute for the individualized modelling case where prior information is available.\\

\FloatBarrier

\paragraph{Stride Length Estimation on Unseen Data}

\begin{figure}[!h]
		\centering
		\includegraphics[width=.95\columnwidth]{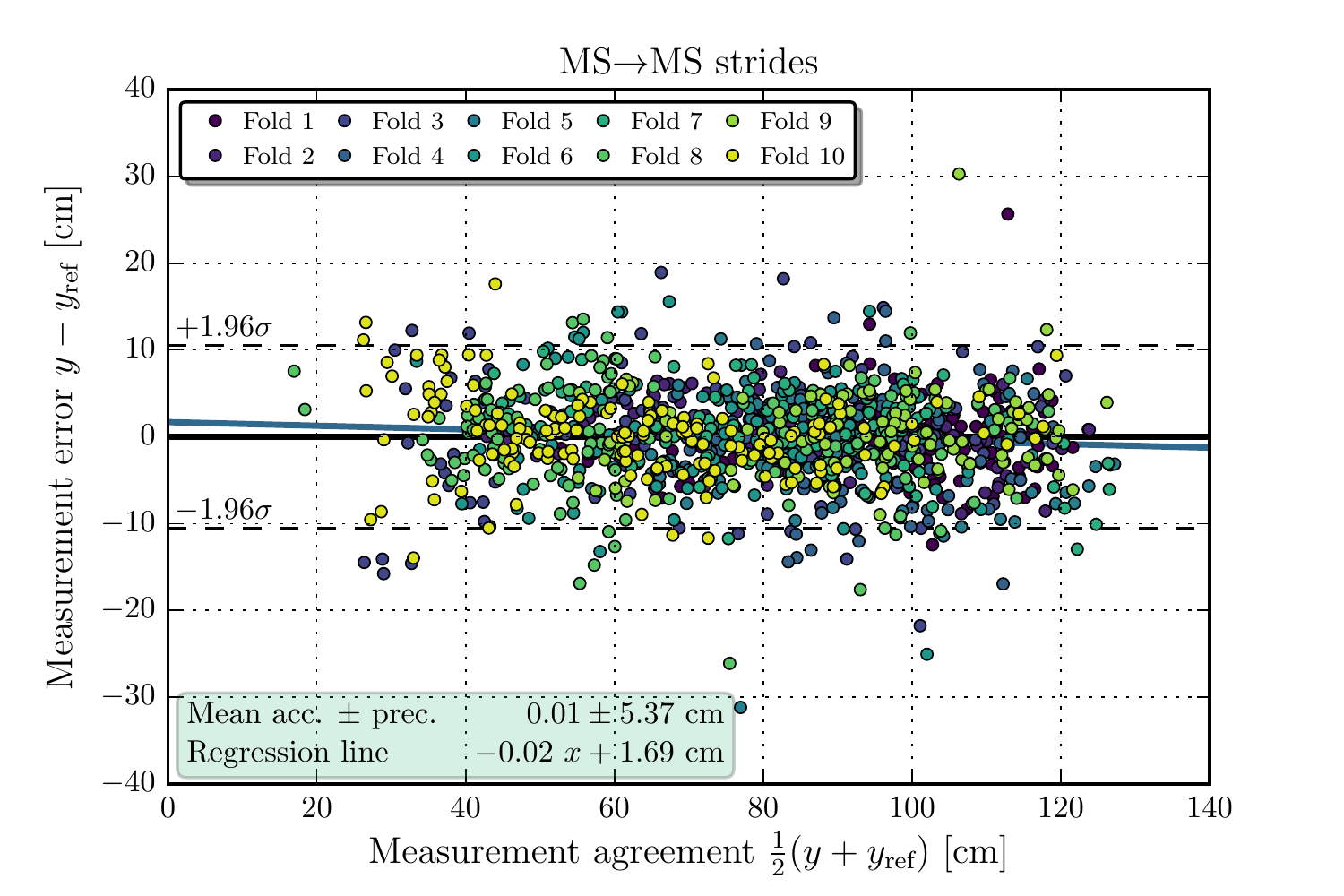}\\\pagebreak
		\includegraphics[width=.95\columnwidth]{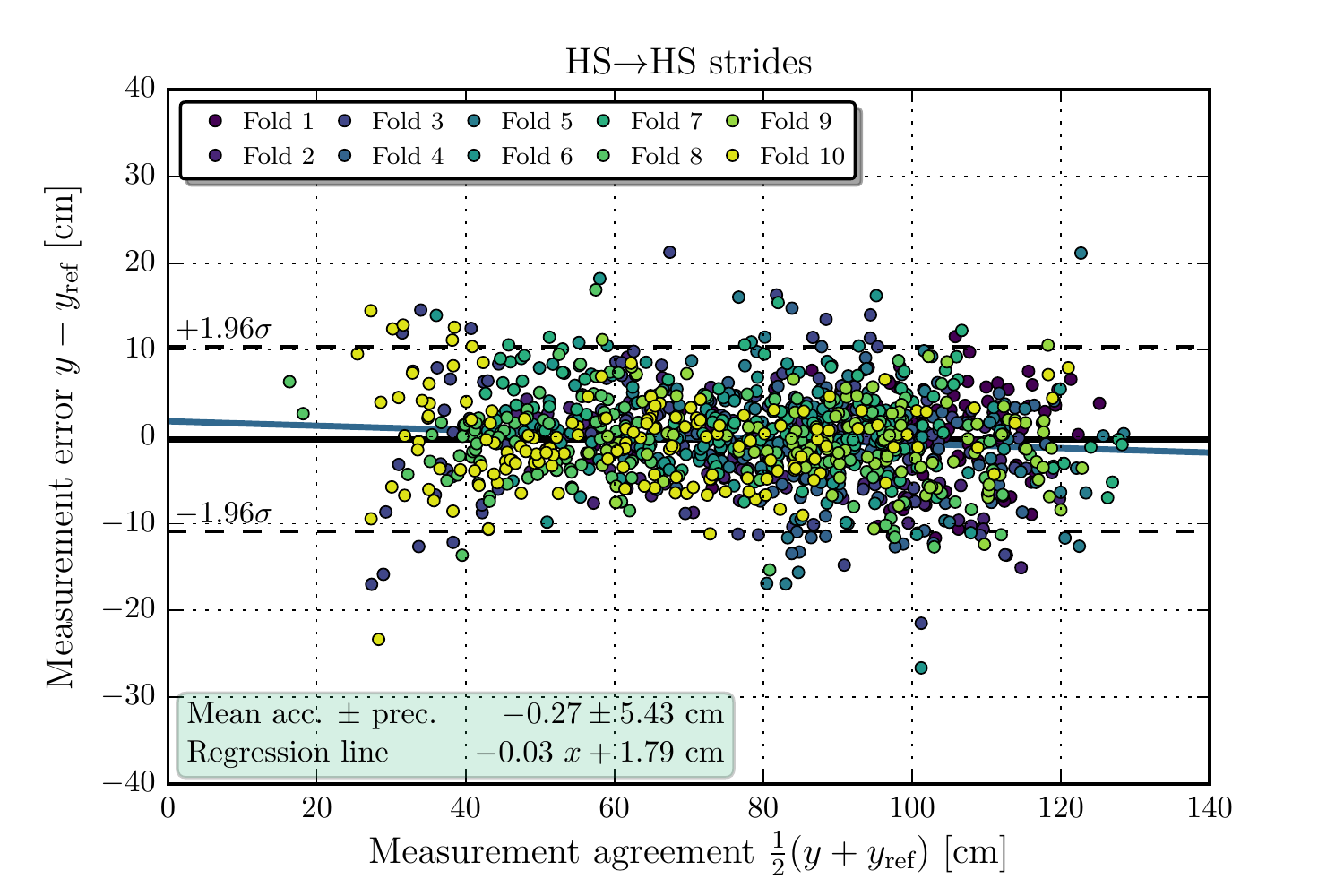}\\
		\includegraphics[width=.95\columnwidth]{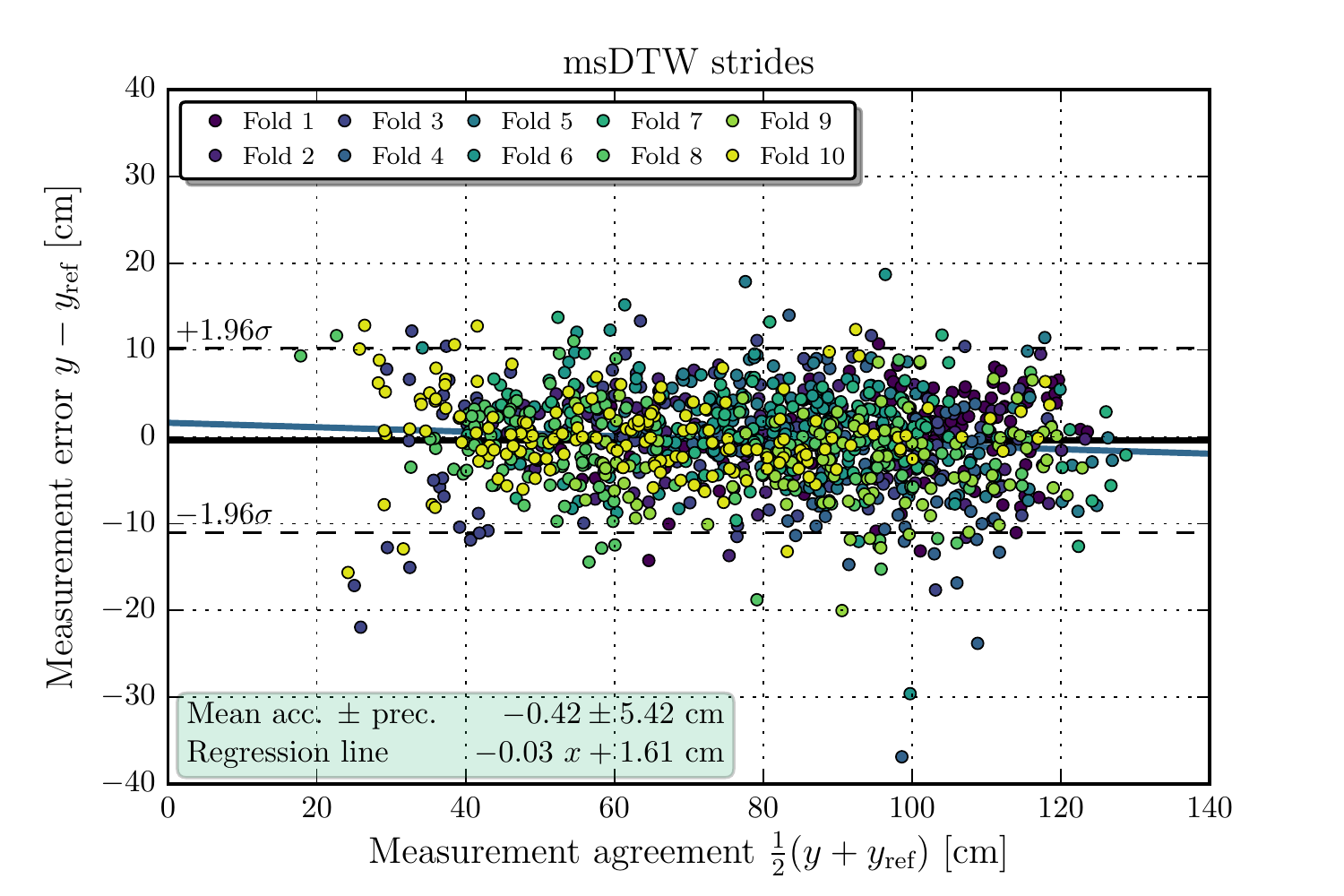}
		\caption{Bland-Altman plots from the evaluation of the proposed algorithm in a 10-fold cross validation for all three stride definitions. Mean accuracy and 95\% confidence intervals ($\pm 1.96~\sigma$) as well as a regression line between mean stride length and the measurement error are shown.}
		\label{fig:bland-altman-plots} 
\end{figure}

Tab. \ref{tab:results} lists average statistics from the 10-fold cross validation for each of the three stride definitions. Bland-Altman plots in Fig. \ref{fig:bland-altman-plots} assess the agreement between the two measurements for all three stride definitions. Additionally, Fig. \ref{fig:bland-altman-plots} includes regression lines between the measurement error $y-y_\text{ref}$ and the measurement agreement $\frac{1}{2} (y+y_\text{ref}) $ in order to assess the dependency of the measurement error w.r.t. the stride length. 

\begin{table*}[!t]
	\centering
	\caption{Results of the deep convolutional neural network approach to stride length estimation for three stride definitions compared to the Gaitrite reference system. Listed are mean stride lengths for both systems, the error of measurement is reported as mean accuracy and precision, relative precision, mean absolute accuracy and precision and Spearman correlation (CC).}
	\begin{tabular}{lccrrcc}
		\toprule
		Stride definition & Our approach & GAITRite &Mean acc. $\pm$ prec. & Rel. prec.  & Mean abs. acc. $\pm$ prec. & CC \\\toprule
		MS$\to$MS strides & $80.03 \pm 22.92$ cm & $80.02 \pm 23.40$ cm & $0.01 \pm 5.37$ cm & $6.7 $ \% & $3.96 \pm 3.63$ cm & $0.97$
\\
		HS$\to$HS strides & $79.75 \pm 22.82$ cm & $80.02 \pm 23.40$ cm & $-0.27 \pm 5.43$ cm & $6.8 $ \% & $4.11 \pm 3.57$ cm & $0.97$
\\
		msDTW strides & $79.60 \pm 22.83$ cm & $80.02 \pm 23.40$ cm & $-0.42 \pm 5.42$ cm & $6.8 $ \% & $4.08 \pm 3.59$ cm & $0.97$
\\
		\bottomrule 		
	\end{tabular}

	\label{tab:results}
\end{table*}
The model yields best performance when strides are defined from MS$\to$MS with a mean accuracy and precision of $0.01\pm 5.37$ cm.
The range of precisions achieved for individual patients ranges between $\pm1.54$ cm in the best case (patient P114) and $\pm10.43$ cm in the worst case (patient P106).

Furthermore, the stride length estimation with the proposed approach is robust w.r.t. the stride definition. When trained and evaluated on the other two stride definitions (HS$\to$HS and msDTW), the mean precision differs only marginally from the one achieved on the MS$\to$MS strides. The mean accuracy, however, decreases to $-0.27$ cm and $-0.42$ cm for HS$\to$HS and msDTW strides respectively. 

The regression lines that indicate the dependency of the measurement error on the stride length regime are comparable for all three stride definitions. The low stride length regime is on average overestimated while high stride lengths are on average underestimated. This reflects an imbalance found in the training data. Since a fixed track length was used for data acquisition, there are more examples from the low stride length regime on the dataset compared to high stride length strides.

\section{Discussion}
Overall, our results show that stride length estimation with deep convolutional neural networks based on stride-specific inertial sensor data is possible. This is the first major contributions of this work. We show feasibility and contribute a completely new type of algorithm to the field of stride length estimation. Although the achieved precision of 5.4 cm is not sufficient to track the subtle yearly decline in stride length in healthy individuals, it is sufficient to resolve stride length differences observed in PD patients \cite{Hass2012}. 

The best result achieved with the method described here outperforms the approach by Rampp et al. \cite{Rampp2015} by $3.0$ cm (36\%) in precision and also gives a more balanced error distribution. We can directly compare these two approaches since they are evaluated on an identical dataset. This is not the case for the other methods found in literature (see Tab. \ref{tab:relatedwork}). Although our approach shows similar performance, the discrepancy in evaluation datasets and their characteristics limits comparability. This manifests the urgent need for a standardised and clinically relevant benchmark dataset regarding stride length estimation from inertial sensor data. Given the number of subjects and the presence of gait alterations in the study population used here, it would serve well as such a benchmark dataset. Furthermore, it is publicly available at \url{https://www5.cs.fau.de/activitynet/benchmark-datasets/digital-biobank/}. 

The second main contribution of this work is the independence of stride definition. The mean accuracy and precision of the approach described is only marginally affected by the stride definition (Tab. \ref{tab:results}). This is particularly interesting since the different events in the datastream forming the stride definition can be determined with different degrees of exactness. The MS event, for example, is defined as the instant of lowest energy in all gyroscope axes during the stride (see \cite{Rampp2015} for details). In cases where the high frequency vibrations caused during heel-strike are not dampened enough to ensure a relatively stable sensor during the stance phase, this can lead to a temporal offset regarding individual MS events. The heel-strike detection, however, is defined as a minimum in the accelerometer x-axis (see \cite{Rampp2015} for details) and can thus be detected more accurately. Subsequently, the MS$\to$MS stride definition is not as exact as for example the HS$\to$HS stride definition. This results in more extreme outliers for the MS$\to$MS strides in the Bland-Altman plots (Fig. \ref{fig:bland-altman-plots}). The fact that this stride definition still yields best mean accuracy and precision is probably coincidence here. 
\FloatBarrier

From a clinical perspective, the independence of stride definition opens up new possibilities regarding analysis of impaired gait. In clinical practice, the zero-velocity assumption is easily violated as in the case of spastic forms of gait alterations as e.g. experienced by post-stroke, multiple sclerosis or spastic paraplegia patients. Standard double-integration methods would fail in estimating spatial stride parameters in these clinically relevant situations. This is not the case with the approach presented here.

Lastly, the high precision on the training dataset shown in Fig. \ref{fig:training} promises excellent results in case of individualization. If we have a training/calibration dataset to tune the neural network to the patient at hand, i.e. by collecting one gait sequence instrumented with inertial sensors and the GAITRite, the proposed method could reach the precision of the reference. This would enable highly accurate stride-by-stride analysis of spatial gait parameters in daily clinical routine or outpatient monitoring at a level sufficient to address clinically relevant phenomena as for example the decline in stride length with disease progression in PD patients or medication effects. Certainly, this has to be formally evaluated in the future and is a possible extension of the presented method.

Furthermore, our methodology is not limited to stride length estimation. Targeting the underlying regression task at a different stride-specific gait characteristic e.g. stride width or timings of individual gait phases does not involve changing the underlying algorithms. The very same technology can be used to estimate different target parameters by simply changing or extending the reference annotation on the training dataset.

The main limitation of the proposed method is that the training of a regression function between stride-specific sensor data and the spatial parameter stride length implicitly depends on the training set. Therefore, the dataset used for training has to capture as much variability as possible for the problem at hand. Otherwise, application on a population that differs from the one used for training might result in lack of model-validity. This is the profound difference between the data-driven approach described here and a double-integration approach that is based on geometric and physical reasoning and does not encounter this kind of problem. However, given enough training data this problem can be solved. Additionally, preliminary experiments already show good generalisation to a PD population without any adaptation of the model trained on the geriatric population presented here. A second profound difference to double-integration approaches is the computational complexity and real-time capability. Although the trained model consists of convolutions and matrix-multiplications with a fixed set of kernels/weights, double-integration approaches will certainly be less complex. Depending on the (medical) application, however, real-time capability might not even be a necessity.

Future work includes a rigorous exploration of the parameter space (number of layers, kernels, weights, etc.) and introduction of a dynamic stop criterion for the optimization that depends on the current performance. Early stopping could also be investigated in this respect.  Additionally, the imbalance on the training data mentioned earlier needs to be addressed. Since limiting the number of strides per patient by random subsampling dramatically reduces the amount of training samples in the current dataset, a balanced training set could for example be acquired by means of simulation. Here, a biomechanical model could be used to sample the space of possible, three-dimensional foot trajectories and output the accelerometer and gyroscope readings for each stride as well as the corresponding target parameter of interest, e.g. stride length. This would also enable much richer datasets for training the network and thereby resolve the aforementioned limitation. Regarding the individualization of the proposed model to a specific patient, this work only shows promising substitute results. This has to be investigated in further detail. Finally, the application of deep convolutional neural networks to other spatial stride parameters, e.g. stride width, or the extraction of complete foot trajectories including foot orientation from stride-specific inertial sensor data will be investigated in future work.

In summary, we used convolutional neural networks to estimate stride length based on stride-specific inertial sensor data captured at the subject's feet. We provide technical validation of the proposed method on a publicly available and clinically relevant benchmark dataset. Furthermore, the proposed approach is robust w.r.t. the stride definition and thus not subject to the zero-velocity phase assumption. Consequently, the current work enables estimation of stride length in clinically relevant situations where the diversity of gait alterations easily violates the assumptions made by state-of-the-art methods.

\FloatBarrier
\section*{Acknowledgements}
This work was supported by the FAU Emerging Fields Initiative (EFIMoves). The authors would like to thank all participants of the study for their contributions.
\bibliographystyle{IEEEtran}
\bibliography{IEEEabrv,NeuralNetRegression_SL}

\end{document}